\documentclass[a4paper,twoside]{article}

\usepackage{algorithm}
\usepackage{algorithmicx}
\usepackage{algpseudocode}
\usepackage[T1]{fontenc}

\usepackage{epsfig}
\usepackage{subcaption}
\usepackage{calc}
\usepackage{amssymb}
\usepackage{amstext}
\usepackage{amsmath}
\usepackage{amsthm}
\usepackage{multicol}
\usepackage{pslatex}
\usepackage{apalike}
\usepackage{SCITEPRESS}     % Please add other packages that you may need BEFORE the SCITEPRESS.sty package.
\usepackage{fancyhdr}

\begin{document}

\title{Edge and Corner Detection in Unorganized Point Clouds for Robotic Pick and Place Applications}

\author{\authorname{Mohit Vohra\sup{1}, Ravi Prakash\sup{1} and Laxmidhar Behera\sup{1,2}}
\affiliation{\sup{1}Department of Electrical Engineering, IIT Kanpur, India}
\affiliation{\sup{2}TCS Innovation Labs, Noida, India}
\email{\{{mvohra, ravipr and lbehera\}@iitk.ac.in}}
}

\keywords{Edge Extraction, Unorganized Point Cloud, Autonomous Grasping}

\pagestyle{plain}
\cfoot{\thepage}

\abstract{In this paper, we propose a novel edge and corner detection algorithm for an unorganized point cloud. Our edge detection method classifies a query point as an edge point by evaluating the distribution of local neighboring points around the query point. The proposed technique has been tested on generic items such as dragons, bunnies, and coffee cups from the Stanford 3D scanning repository. The proposed technique can be directly applied to real and unprocessed point cloud data of random clutter of objects. To demonstrate the proposed technique's efficacy, we compare it to the other solutions for 3D edge extractions in an unorganized point cloud data. We observed that the proposed method could handle the raw and noisy data with little variations in parameters compared to other methods. We also extend the algorithm to estimate the 6D pose of known objects in the presence of dense clutter while handling multiple instances of the object. The overall approach is tested for a warehouse application, where an actual UR5 robot manipulator is used for robotic pick and place operations in an autonomous mode.}

\onecolumn \maketitle \normalsize \setcounter{footnote}{0} \vfill

\section{Introduction}
Over the past decade, the construction industry has struggled to improve its productivity, while the manufacturing industry has experienced a dramatic productivity increase \cite{changali2015construction} \cite{vohra2019real} \cite{pharswan2019domain}. A possible reason is the lack of advanced automation in construction \cite{asadi2018real}. However, recently various industries have shown their interest in construction automation due to the following benefits, i.e., the construction work will be continuous, and as a result, the construction period will decrease, which will provide tremendous economic benefits. Additionally, construction automation improves worker's safety and enhances the quality of work. The most crucial part of the construction is to build a wall and to develop a system for such a task; the system should be able to estimate the brick pose in the clutter of bricks, grasp it, and place it in a given pattern. Recently, a New York-based company, namely "Construction Robotics", has developed a bricklaying robot called SAM100 (semi-automated mason) \cite{parkes2019automated}, which makes a wall six times faster than a human. The SAM100 requires a systematic stack of bricks at regular intervals, making this system semi-autonomous, as the name suggests.

\begin{figure}[t!]
\centering
  \includegraphics[height=4cm]{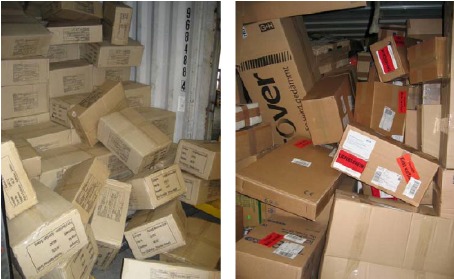}
  \caption{Cartons Clutter (Image Courtesy: Internet)}
  \label{fig:cartons_clutter}
\end{figure}

Similarly, in the warehouse industry, the task of unloading goods from trucks or containers is crucial. With the development of technology, various solutions have been proposed to incorporate automation in unloading goods \cite{doliotis20163d} \cite{stoyanov2016no}. One of the main challenges in the automation of the above work is that the system has to deal with a stack of objects, which can come in random configurations, as shown in Fig. \ref{fig:cartons_clutter}. For developing a system for such a task, the system must estimate the pose of the cartons in a clutter, grasp it, and arrange it in the appropriate stack for further processing.

In this paper, we focus on estimating the pose of the objects (carton or brick) in clutter, as shown in Fig. \ref{fig:cartons_clutter}. We assume that all the objects present in the clutter have the same dimensions, which is very common in warehouse industries (clutter of cartons all having the same dimensions) and construction sites (clutter of bricks). Traditionally, Object pose is estimated by matching the 2D object features \cite{lowe2004distinctive} \cite{collet2011moped} or 3D object features \cite{drost2010model} \cite{hinterstoisser2016going} between object model points and view points. Recently, CNN based solutions have shown excellent results for 6D pose estimation, which exempts us from using hand-crafted features. For example, \cite{xiang2017posecnn} \cite{tekin2018real} can estimate the pose of the objects directly from raw images. Similarly, \cite{qi2017pointnet}\cite{zhou2018voxelnet} can directly process the raw point cloud data and predict the pose of the objects. While all the above methods have shown excellent results, they cannot be used directly for our work.

\begin{itemize}
    \item Since the objects are textureless. Therefore the number of features will be less, making the feature matching methods less reliable. Furthermore, due to the presence of multiple instances of the object, it can be challenging to select features that correspond to a single object.
    \item  The performance of CNN-based algorithms relies on extensive training on large datasets. However, due to the large variety of objects in the warehouse industry (various sizes and colors), it is challenging to train CNN for all types of objects. 
\end{itemize}

Therefore, we aim to develop a method that requires fewer features and can easily be deployed for objects with different dimensions.

The main idea of our approach is that if we can determine at least three corner points of the cartons, then this information is sufficient to estimate its pose. So the first step is to extract sharp features like edges and boundaries in point cloud data. In the case of edge detection in unstructured 3D point clouds, traditional methods reconstruct a surface to create a mesh \cite{kazhdan2013screened} or a graph \cite{demarsin2007detection} to analyze the neighborhood of each point. However, reconstruction tends to wash away sharp edges, and graph-based methods are computationally expensive. In \cite{bazazian2015fast}, the author performs fast computation of edges by constructing covariance matrices from local neighbors, but the author has demonstrated for synthetic data. In \cite{ni2016edge}, the authors locate edge points by fitting planes to local neighborhood points. Further a discriminative learning based approach \cite{hackel2016contour} is applied to unstructured point clouds. The authors train a random forest-based binary classifier on a set of features that learn to classify points into edge vs. non-edge points. One drawback of their method is poor performance on unseen data.

In this paper, we present a simple method to extract edges in real unstructured point cloud data. In this approach, we assign a score to each point based on its local neighborhood distribution. Based on the score, we classify a point as an edge (or boundary) point or a non-edge point. From the edge points, we find the line equation, the length of the 3D edges, and the corner points, which are intersections of two or more edges. For each corner point, the corresponding points in the local object frame are calculated to estimate the object's 6D pose. Our approach's main advantage is that it can be easily applied to other objects with known dimensions.

The remainder of this paper is organised as follows. Proposed 3D edge extraction is presented in Section II. In Section III corner points are found using those edges. Pose estimation using edges and corners are presented in Section IV. In Section V, experimental results for robotic manipulation using a 6-DOF robot manipulator and their qualitative comparison with the state-of-the-art is done. This paper is finally concluded
in Section VI.

\section{Edge Points extraction}

In this section, we will explain the method for extracting the edge points. Input to the algorithm is raw point cloud data $\mathcal{P}$, user-defined radius $r_s$, and threshold value $t_h$. The output of the algorithm is point cloud $\mathcal{E}$ which contains the edge points. To decide if a given query point $\mathbf{p_{i}} \in \mathcal{P}$ is an edge point or not, we calculate $\mathcal{R}(\mathbf{p_{i}})$, which is defined as a set of all points inside a sphere, centered at point $\mathbf{p_{i}}$ with radius $r_s$. For an unorganized point cloud, this is achieved through a k-dimensional (K-d) tree. We call each point in set $\mathcal{R}(\mathbf{p_{i}})$ as a neighboring point of $\mathbf{p_{i}}$, and it can be represented as $\mathcal{R}(\mathbf{p_{i}}) = \begin{Bmatrix}
\mathbf{n_{1}, n_{2}, \dots , n_{k}}
\end{Bmatrix}$. For each query point $\mathbf{p_{i}}$ and neighboring point $\mathbf{n_{j}}$ we calculate the directional vector as

\vspace{-5mm}

\begin{align}
\label{eqn:eq1}
 \mathbf{d(p_{i}, n_{j}) = n_{j} - p_{i}}\\
 \mathbf{\hat{d}(p_{i}, n_{j})} = \frac{\mathbf{d(p_{i}, n_{j})}}{\left \| \mathbf{d(p_{i}, n_{j})} \right \|}
\end{align}

Then we calculate the resultant directional vector $\mathbf{\hat{R}(p_{i})}$ as sum of all directional vector and normalize it
\begin{align}
% \label{eqn:eq2}
 \mathbf{R(p_{i})} = \sum_{j=1}^{k}\mathbf{\hat{d}(p_{i}, n_{j})},
 \mathbf{\hat{R}(p_{i})} = \frac{\mathbf{R(p_{i})}}{\left \|  \mathbf{R(p_{i})}  \right \|}
\end{align}

We assign a score $s(\mathbf{p_{i}})$ to each query point $\mathbf{p_{i}}$ as an average of the dot product between $\mathbf{\hat{R}(p_{i})}$ and $\mathbf{\hat{d}(p_{i}, n_{j})}$ for all neighboring points.
\begin{align}
\label{eqn:eq3}
 s(\mathbf{p_{i}}) = \frac{\sum_{j=1}^{k}\mathbf{\hat{R}(p_{i})} \cdot \mathbf{\hat{d}(p_{i}, n_{j})}}{k}
\end{align}

\begin{figure}[h!]
  \centering
  \begin{subfigure}[b]{0.4\linewidth}
    \includegraphics[width=\linewidth]{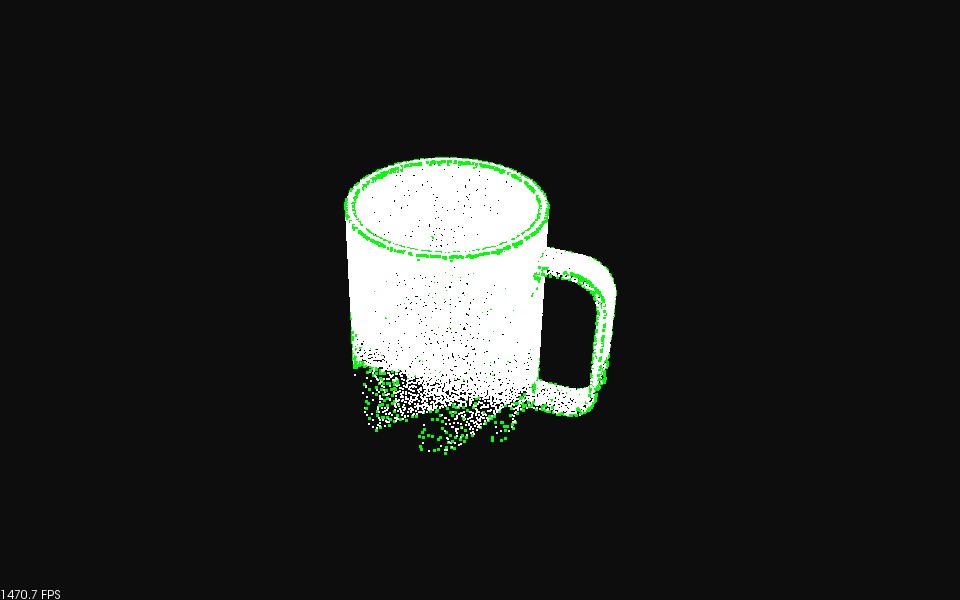}
    \caption{Coffee mug}
  \end{subfigure}
    \begin{subfigure}[b]{0.4\linewidth}
    \includegraphics[width=\linewidth]{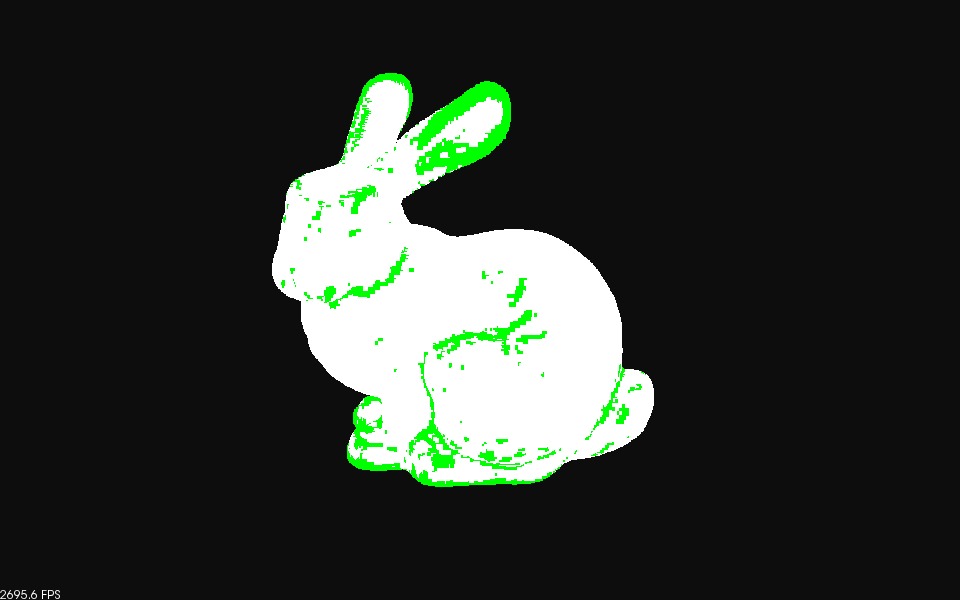}
    \caption{bunny}
  \end{subfigure}
  \begin{subfigure}[b]{0.4\linewidth}
    \includegraphics[width=\linewidth]{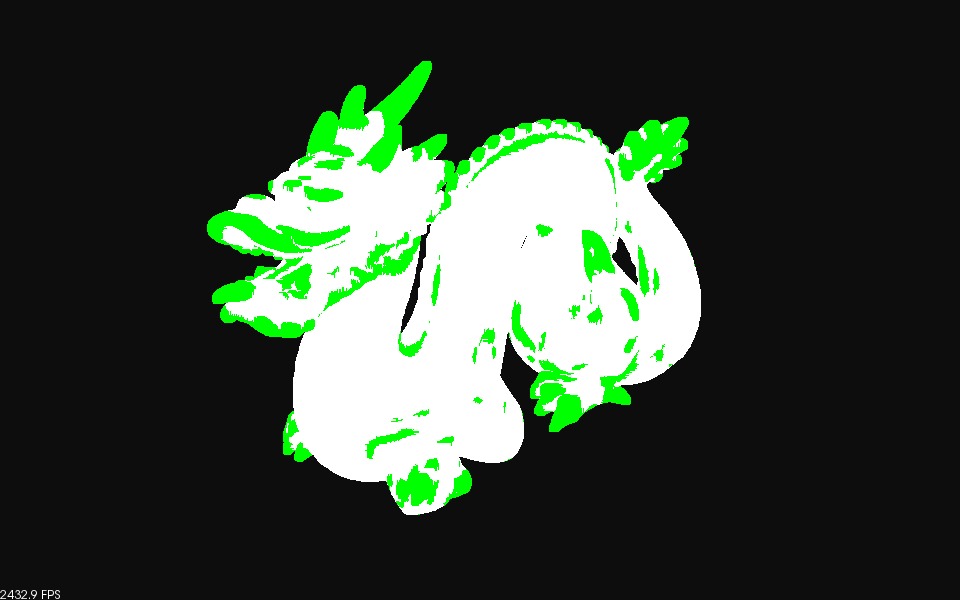}
    \caption{dragon}
  \end{subfigure}
    \begin{subfigure}[b]{0.4\linewidth}
    \includegraphics[width=\linewidth]{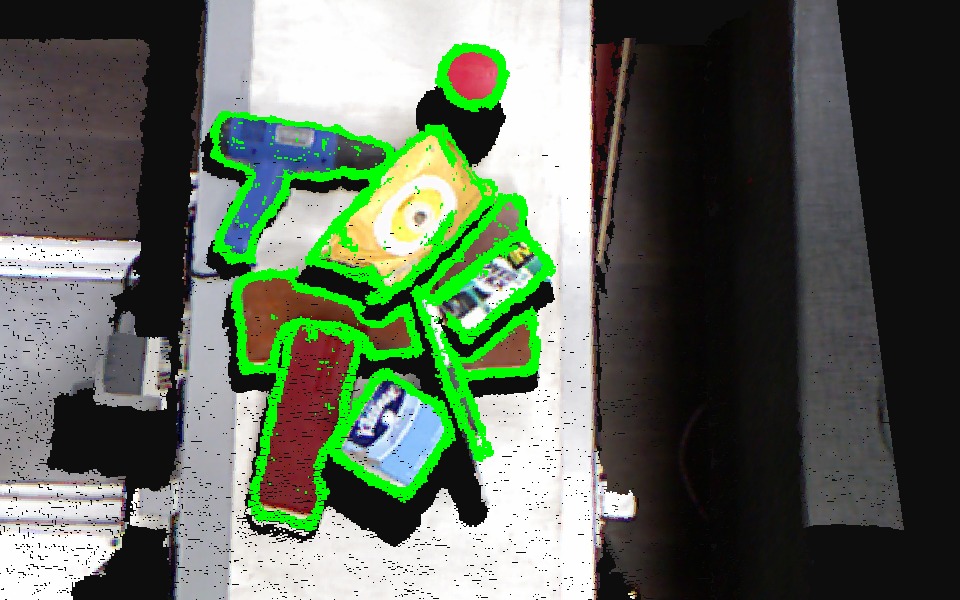}
    \caption{Objects Clutter}
  \end{subfigure}
  
  \caption{Edge cloud}
  \label{fig:edge cloud}
\end{figure}

If $s(\mathbf{p_{i}})$ exceeds some threshold $t_h$, $\mathbf{p_{i}}$ is considered an edge point, otherwise not. We have tested the algorithm on standard data set from Stanford 3D Scanning Repository\footnote{http://graphics.stanford.edu/data/3Dscanrep/} and \cite{lai2014unsupervised}, with point cloud models of various objects such as bowls, cups, cereal boxes, coffee mugs, soda cans, bunnies, and dragons. We have also applied our algorithm to a clutter of random objects to extract edges, shown in Fig. \ref{fig:edge cloud}.

 \vspace{-3mm}

\section{Estimating the Length of Edges}
Once we have found all the edge points, our next step is to find the line equation of edges, length of edges, and corner points. In general, the equation of a line is represented by two entities, a point through which line passes and a unit directional vector from that point. However, in our case, we represent the line with two extreme points. The advantage of the above representation is that we can estimate the length of the edge by calculating the distance between two extreme points. The intersecting point between two edges can be found by calculating the distance between the extremities of edges.

\subsection{Reference index and line Equation}
To find the line equation, we apply the \textit{RANSAC} method on the edge point cloud $\mathcal{E}$. The output of the \textit{RANSAC} method is $\mathcal{L}$ and $\mathcal{I}$, where $\mathcal{L}$ represents the equation of a line in standard form i.e., a point and a directional vector, and $\mathcal{I}$ represent the set of inliers. To find two extremities of line $\mathcal{L}$, we have to make sure that two extremities should represent the same edge because, in a cluttered environment, $\mathcal{I}$ could have points that belong to different objects. So for this task, we will find the reference point $\mathbf{p_{r}}$. We define a point $\mathbf{p_{r}} \in \mathcal{I}$, which has the maximum number of neighbors in $\mathcal{I}$, and $r$ is the reference index of $\mathbf{p_{r}}$ in $\mathcal{I}$. The maximum number of neighbors property ensures that $\mathbf{p_{r}}$ is not at the extreme ends, and from $\mathbf{p_{r}}$ we will find the two extreme points, which is explained in the next section. Following are the steps to estimate the reference index $r$
\begin{itemize}
    \item  Use \textit{RANSAC} on $\mathcal{E}$ for estimating the line equation $\mathcal{L}$ and set of inliers $\mathcal{I}$.
    \item We define $\mathcal{R}(\mathbf{p_{i}})$ as set of all inlier points inside a sphere, centered at point $\mathbf{p_{i}}$, where $\mathbf{p_{i}} \in \mathcal{I}$
    \item we define $ r  = \underset{i} {\mathrm{argmax}} ~size(\mathcal{R}(\mathbf{p_{i}}))$
\end{itemize}

\subsection{Finding extremities}
In this section, we will explain the procedure to find the extreme points $\mathbf{e_{1}}$ and $\mathbf{e_{2}}$ in a set of points $\mathcal{I}$ for a reference point $\mathbf{p_{r}}$. To test if a query point $\mathbf{p_{i}} \in \mathcal{I}$  is an extreme point or not, we find a set of inliers $\mathcal{R(\mathbf{p_{i}})}$, with radius $r_s$ and represent the set as $\mathcal{R}(\mathbf{p_{i}}) = \begin{Bmatrix}
\mathbf{n_{1}, n_{2}, \dots , n_{k}}
\end{Bmatrix}$. For each query point $\mathbf{p_{i}}$, we calculate the directional vector from a query point to the reference point ${\mathbf{d_1}} = \mathbf{p_r} - \mathbf{p_i}$, and a local directional vector from the query point to the neighbor points ${\mathbf{d_2}} = \mathbf{n_j} - \mathbf{p_i}$, where $\mathbf{n_j} \in \mathcal{R(\mathbf{p_{i}})}$. Now to decide if the query point is an extreme point or not, we compute the dot product between ${\mathbf{d_1}}$ and ${\mathbf{d_2}}$ for all neighboring points. If any of the dot product value is negative, then $\mathbf{p_i}$  is not an extreme point; otherwise, it is. We repeat the procedure for all $\mathbf{p_i}$  and select the extreme points which have the smallest distance from $\mathbf{p_r}$. Implementation of the above procedure is given in \textit{Algorithm \ref{algo_extremas}}.

\begin{algorithm}
  \caption{Get extreme points}
  \label{algo_extremas}
  \begin{algorithmic}[1]
  \Require $r$, $\mathcal{I}$ 
  \Ensure{$\mathbf{e_1}, \mathbf{e_2}$}  \Comment{extreme points}
  \For{$i = 0 \to size(\mathcal{I})$ and $i \neq r$} 
    \State find $\mathcal{R}(\mathbf{p_{i}})$
    \State ${\mathbf{d_1}} = \mathbf{p_{r}} - \mathbf{p_i}$
    \State $\hat{\mathbf{d_1}} = \frac{\mathbf{d_1}}{\left \| \mathbf{d_1} \right \| }$  
    
    \For{$j = 0 \to size(\mathcal{R}(\mathbf{p_{i}}))$} 
       \State ${\mathbf{d_2}} = \mathbf{n_j} - \mathbf{p_i}$
       \State $\hat{\mathbf{d_2}} = \frac{\mathbf{d_2}}{\left \| \mathbf{d_2} \right \| }$
       \State $v = \hat{\mathbf{d_1}} \cdot \hat{\mathbf{d_2}}$ 
      \If{ $v < 0$}  \Comment{$\mathbf{p_{i}}$ is not an extreme point}  
          \State break
      \EndIf
    \EndFor
          \If{ $v > 0$}  \Comment{$\mathbf{p_{i}}$ can be an extreme point}
          \State ${\mathbf{p_i}}  \to points$ \Comment{save point}
          \State $\hat{\mathbf{d_1}}  \to directions$ \Comment{save direction}
          \State $\left \| \mathbf{d_1} \right \|  \to distances$ \Comment{save distance}
      \EndIf
    
  \EndFor
  
  \State $\mathbf{e_1} \to min dis (distances)$
  \State $\mathbf{e_2}$ is first minima opposite to $\mathbf{e_1}$ \Comment{using saved $\hat{\mathbf{d_1}} $}
  \end{algorithmic}
\end{algorithm}

\subsection{Get all edges}
To find all the lines in the edges point cloud $\mathcal{E}$, we recursively apply the above algorithms, and the points corresponding to the line will be removed from $\mathcal{E}$ to obtain the new line equation. The following is a sequence of steps to extract all the lines.
\begin{itemize}
    \item  Let $\mathcal{P}$ be the raw Point cloud data, $r_s$ be the radius and $t_h$ be the threshold value to extract the edges point cloud $\mathcal{E}$.
    \item Apply \textit{RANSAC} on $\mathcal{E}$ to get reference index $r$ and set of inliers $\mathcal{I}$.
    \item Extract two extreme points $\mathbf{e_1}$ and $\mathbf{e_2}$.
    \item Store $\mathbf{e_1}$ and $\mathbf{e_2}$ in an array $\mathbf{A}$ and remove all the points between $\mathbf{e_1}$ and $\mathbf{e_2}$ from $\mathcal{E}$.
    \item repeat the steps $2 - 4$.
\end{itemize}

\section{Pose estimation}
\subsection{Club edges of same cube}

\begin{algorithm}
  \caption{Get edges of same cube}
  \label{algo_edges_of_same_cube}
  \begin{algorithmic}[1]
  \Require $\mathbf{A}$, $a$
  \Ensure $\mathbf{C}$ \Comment{all edges of a cube}
  
  \State $\mathbf{l_{1}} = \mathbf{A}(\textit{a})$
  
  \While{not traverse all edges}
  \State $\mathbf{l_{2}} \in \mathbf{A} - \mathbf{l_{1}}$
  \If{$\mathbf{l_{1}} \perp \mathbf{l_{2}}$ and $\mathbf{l_{1}} \cap \mathbf{l_{2}} \neq \phi$}
      \State $\mathbf{l_{1}}, \mathbf{l_{2}} \to \mathbf{C}$     \Comment{save $\mathbf{l_{1}}$, $\mathbf{l_{2}}$ in $\mathbf{C}$}
      
    \State $\mathbf{A} \not\equiv \mathbf{l_{1}}$   \Comment{remove $\mathbf{l_{1}}$ from $\mathbf{A}$}      
    
    \State $\mathbf{l_{1}} = \mathbf{l_{2}}$   \Comment{replace $\mathbf{l_{1}}$ with $\mathbf{l_{2}}$}

  \EndIf
  \EndWhile
  
%   \Ensure Remove Duplicates from $\mathbf{C}$

  \end{algorithmic}
\end{algorithm}

Once we have all the edges in $\mathbf{A}$, our next step is to club the edges of the same cube. Let \textit{a} be an edge number, and the corresponding edge is $\mathbf{l_{1}} = \mathbf{A}(\textit{a})$. If $\mathbf{l_{1}}$ represents one edge of a cube, then we need to find all other edges of the same cube. To find other edges, we consider the property that $\mathbf{l_{2}}$ must be orthogonal to $\mathbf{l_{1}}$, i.e., $\mathbf{l_{1}} \perp \mathbf{l_{2}}$, and intersect with each other at a point, i.e., $\mathbf{l_{1}} \cap \mathbf{l_{2}} \neq \phi $. If $\mathbf{l_{2}}$ satisfies the above conditions, we store $\mathbf{l_{1}}$ and $\mathbf{l_{2}}$ in an array and replace $\mathbf{l_{1}}$ with $\mathbf{l_{2}}$. We repeat the above steps until we cover all the edges. The process is described in \textit{Algorithm \ref{algo_edges_of_same_cube}}. Once we have all the edges representing the same cube, we compute the corner points, which are intersections of two or more edges. These corner points are calculated in the camera frame, and by estimating the corresponding points in the local object frame, we can estimate the object's pose.

\begin{figure}[ht]
\centering
  \begin{subfigure}[b]{0.4\linewidth}
    \includegraphics[width=\linewidth]{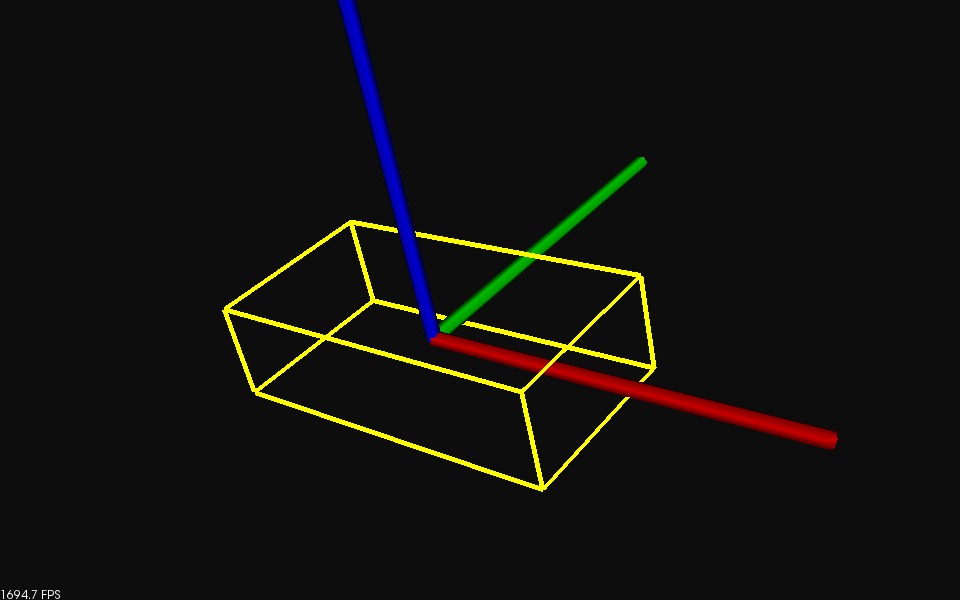}
    \caption{Local Brick frame}
  \end{subfigure}
    \begin{subfigure}[b]{0.4\linewidth}
    \includegraphics[width=\linewidth]{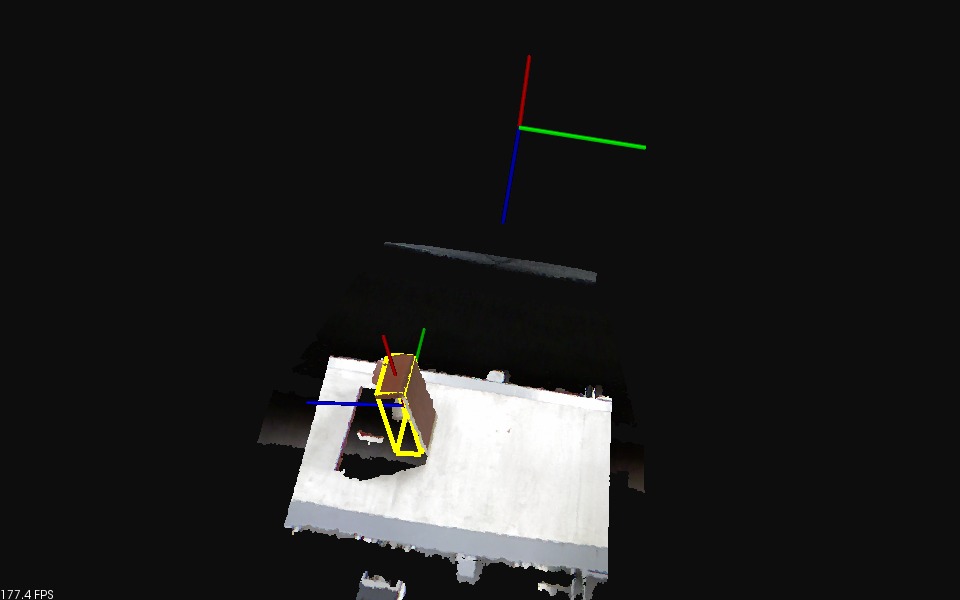}
    \caption{Pose using \textit{b}, \textit{h}}
  \end{subfigure}
    \begin{subfigure}[b]{0.4\linewidth}
    \includegraphics[width=\linewidth]{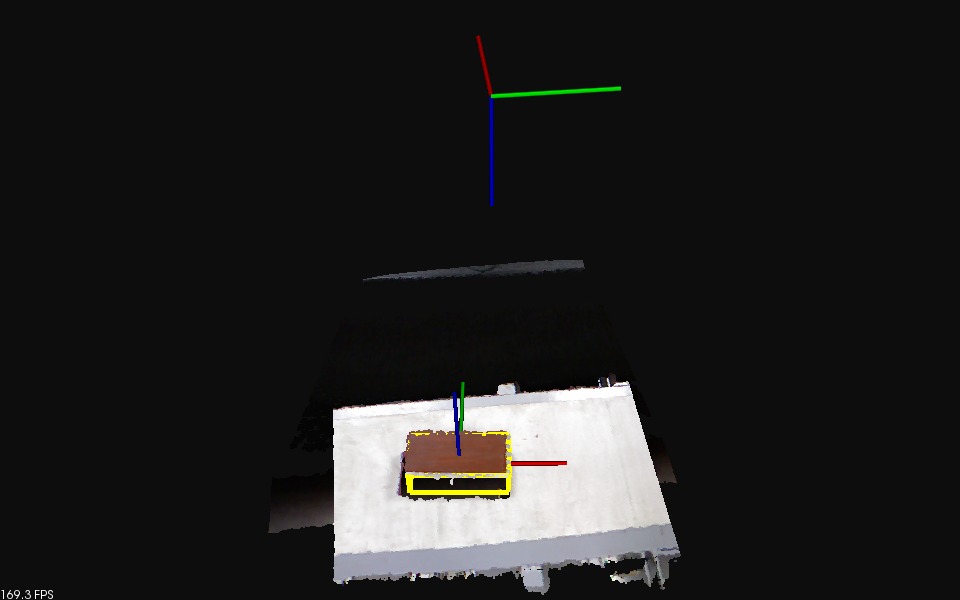}
    \caption{Pose using \textit{l}, \textit{b}}
  \end{subfigure}
    \begin{subfigure}[b]{0.4\linewidth}
    \includegraphics[width=\linewidth]{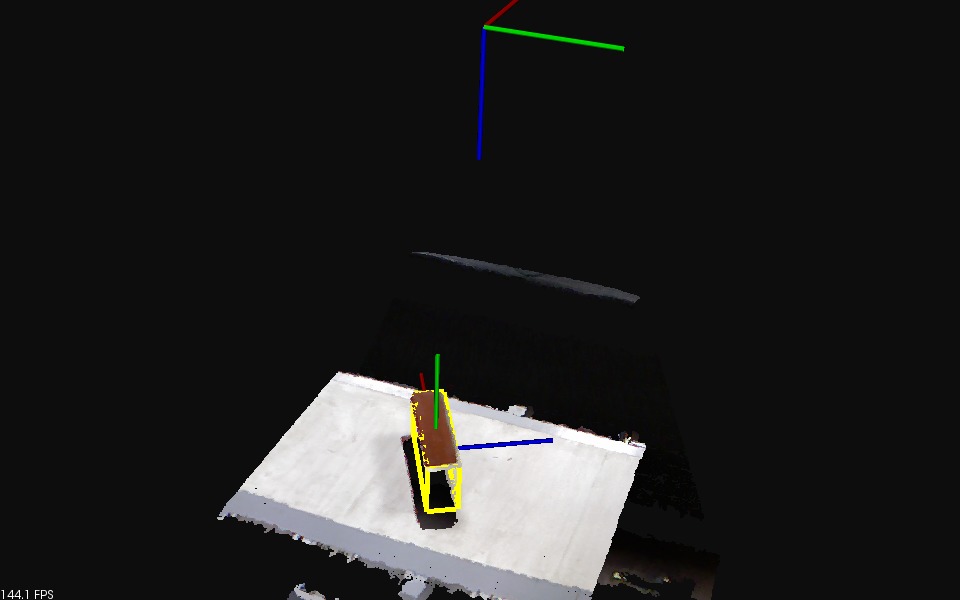}
    \caption{Pose using \textit{l}, \textit{h}}
  \end{subfigure}
  \caption{Local brick frame and assumptions}
  \label{fig:Local cube frame and Assumptions}
\end{figure}

\subsection{Corresponding points and Pose estimation}

\subsubsection{Local frame and assumptions}
Let \textit{l}, \textit{b}, \textit{h} represent the length, breadth, and height of the cube, and we define a local frame with the origin at the centroid of the cube, and x-axis, y-axis, and z-axis pointing along the length, breadth, and height of the cube respectively, shown in Fig. \ref{fig:Local cube frame and Assumptions}. 

For estimating the pose, we need a minimum of two orthogonal edges. Although two orthogonal edges can give two poses, we will assume that the third axis corresponding to the remaining edge will point to camera origin. So this assumption filter out the one solution, and we will get a unique pose from two edges shown in Fig. \ref{fig:Local cube frame and Assumptions}.

\subsubsection{Corner points and directional vector}
Let two edges are $\mathbf{l_1}$, $\mathbf{l_2}$, and from these edges, we will calculate three corner points, $\mathbf{p_1}$, $\mathbf{p_2}$, $\mathbf{p_3}$, where $\mathbf{p_1}$ is the intersecting point of two lines, $\mathbf{p_2}$ is another extreme of $\mathbf{l_1}$ and $\mathbf{p_3}$ is another extreme of $\mathbf{l_2}$. We also define four unit directional vector $\mathbf{d_{1}}, \mathbf{d_{2}}, \mathbf{t},$ and $\mathbf{d}$ as

\begin{align}
\label{eqn:eq4}
    \mathbf{d_{1}} =  \frac{\mathbf{p_{2}} - \mathbf{p_{1}}}{\left \| \mathbf{p_{2}} - \mathbf{p_{1}} \right \|},\;
  \mathbf{d_{2}} =  \frac{\mathbf{p_{3}} - \mathbf{p_{1}}}{\left \| \mathbf{p_{3}} - \mathbf{p_{1}} \right \|},\;
  \mathbf{t} =  \frac{-\mathbf{p_{1}}}{\left \|\mathbf{p_{1}} \right \|}
\end{align}
\vspace{-1em}
\begin{align}
\label{eqn:eq5} 
  \mathbf{d} = \mathbf{d_{1}} \times  \mathbf{d_{2}},\;
  \mathbf{d} =  \frac{\mathbf{d}}{\left \|\mathbf{d} \right \|}
\end{align}
\vspace{-2mm}

Vector $\mathbf{d_{1}}$ represents the unit directional vector corresponding to edge $\mathbf{l_{1}}$, $\mathbf{d_{2}}$ represents the unit directional vector corresponding to edge $\mathbf{l_{2}}$ and $\mathbf{t}$ represents the unit directional vector from $\mathbf{p_1}$ to camera origin. Vector $\mathbf{d}$ is a unit directional vector orthogonal to $\mathbf{d_1}$ and $\mathbf{d_2}$.

\subsubsection{Direction assignment and corresponding points}
Once we have the corner points, our next step is to find their corresponding points in the local object frame. For estimating the corresponding points, it is important to verify if the $\mathbf{d_{1}}$( or $\mathbf{d_{2}}$) is representing the +x(y,z) or -x(y,z) wrt local object frame. Further, we have to assign the direction to $\mathbf{d_{1}}$ and $\mathbf{d_{2}}$ in such a way that third remaining axis of the brick should point towards the camera frame.

\begin{figure}
  \includegraphics[width=\linewidth]{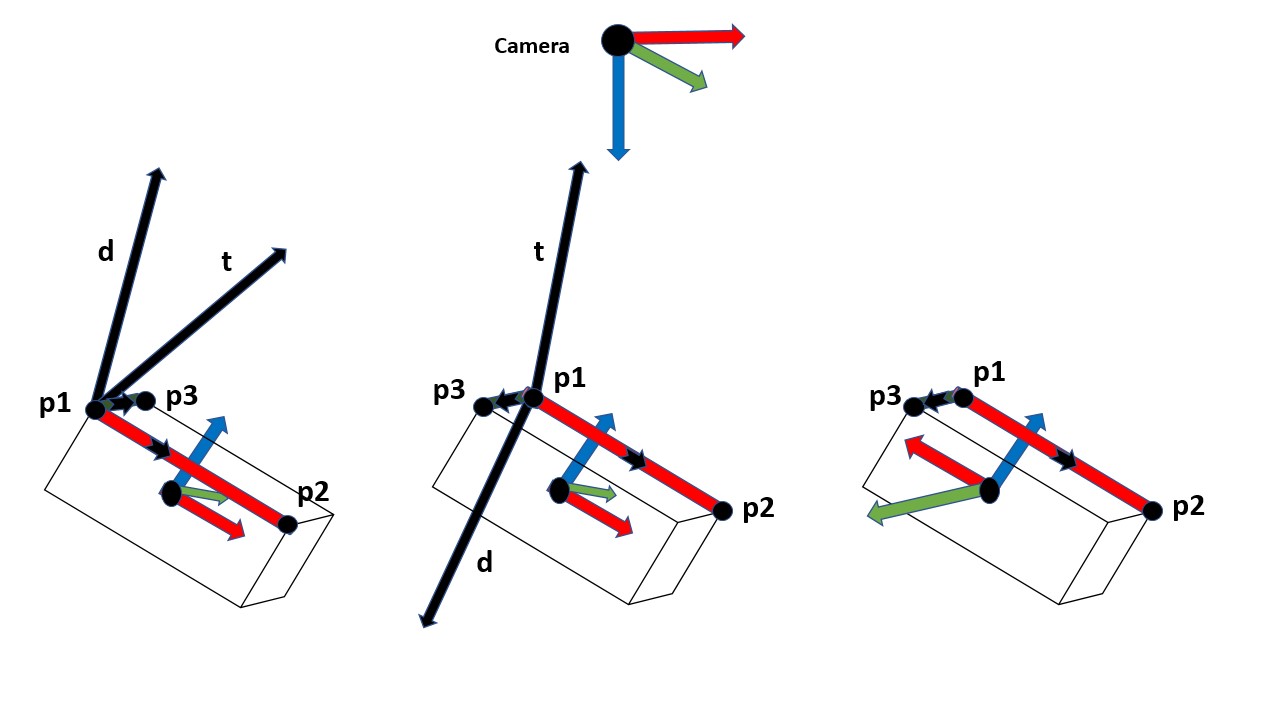}
  \caption{Direction assignment and corresponding points}
  \label{fig:direction}
\end{figure}

We explain direction assignment with an example. Let $\left \| \mathbf{l_{1}} \right \| = l$ and $\left \| \mathbf{l_{2}} \right \| = b$. Since remaining edge is the height edge so corresponding axis i.e. z-axis of local frame must point towards the camera frame which is shown in Fig. \ref{fig:direction}. From the two edges we compute the corner points and directional vectors. For visualization, we represent $\mathbf{l_{1}}$ in red color because $\mathbf{l_{1}}$ must have a direction of either $\hat{x}$ or $-\hat{x}$ in local frame. Similarly we represent $\mathbf{l_{2}}$ in green color because $\mathbf{l_{2}}$ must have a direction of either $\hat{y}$ or $-\hat{y}$ in local frame. Directional vector and corner points are represented in black color (Fig. \ref{fig:direction}). Let us assume that $\mathbf{d_{1}}$ has a direction of $\hat{x}$ and $\mathbf{d_{2}}$ has a direction of $\hat{y}$ in local frame. To verify our assumption, we compute $\mathbf{d}$ which represents the directional vector along the third axis (z-axis) and validate our assumptions by computing the dot product between $\mathbf{d}$ and $\mathbf{t}$.
\par
As shown in Fig. \ref{fig:direction}, for the first brick our direction assumptions are correct because dot product between $\mathbf{d}$ and $\mathbf{t}$ is positive and corresponding points are given in Table \ref{correspomdences}.
\par
For second brick, our direction assumptions are incorrect because dot product between $\mathbf{d}$ and $\mathbf{t}$ is negative, which means either $\mathbf{d_{1}}$ has a direction of $-\hat{x}$ or $\mathbf{d_{2}}$ has a direction of $-\hat{y}$. One can conclude from second brick (Fig. \ref{fig:direction}), that $\mathbf{d_{1}}$ with a direction of $\hat{x}$ and $\mathbf{d_{2}}$ with a direction of $-\hat{y}$ is the valid solution and corresponding points are given in Table\ref{correspomdences}.

\begin{table}[ht]
	\fontsize{7pt}{11pt}\selectfont
	\centering
	\caption{Correspondences}
	\label{correspomdences}
	\begin{tabular}{|c|c|c|}
	    \hline
		\textbf{Corner points}& \textbf{Brick } $\mathbf{1}$ & \textbf{Brick } $\mathbf{2}$\\
		\hline
		$\mathbf{p_{1}}$ & $-\frac{l}{2}, -\frac{b}{2}, \frac{h}{2}$ & $-\frac{l}{2}, \frac{b}{2}, \frac{h}{2}$ \\
		\hline
		$\mathbf{p_{2}}$ & $\frac{l}{2}, \frac{b}{2}, \frac{h}{2}$ & $\frac{l}{2}, \frac{b}{2}, \frac{h}{2}$ \\
		\hline
		$\mathbf{p_{3}}$ & $-\frac{l}{2}, -\frac{b}{2}, \frac{h}{2}$ & $ -\frac{l}{2}, -\frac{b}{2},  \frac{h}{2}$ \\
		\hline
	\end{tabular}
\end{table}

If we consider the other solution i.e. $-\hat{x}$ for $\mathbf{d_{1}}$ and $\hat{y}$ for $\mathbf{d_{2}}$, then as shown in third brick, local cube frame rotates by an angle $180$ about z-axis, and since the cube is symmetric, so both solutions will project the same model. 

\subsubsection{Pose refinement}
Using three correspondences, we compute the pose of the cube. To further refine the pose, we need more correspondences. From \textit{Algorithm \ref{algo_edges_of_same_cube}} we get $\mathbf{C}$ which represents all edges of same cube and intersection of edges will give the corner points. From the initial pose calculated from three correspondences, we will find the correspondence between the other corner points in camera frame and in local frame by calculating the distance between corner points and the predicted corner points from the initial pose. Pair of corner point and predicted corner point with minimum distance will form a correspondence pair. Thus more correspondences will give more accurate pose.

\section{Experimental Results and Discussion}
\begin{figure*}
 \begin{subfigure}[b]{0.2\textwidth}
                \centering
                \includegraphics[scale=0.1]{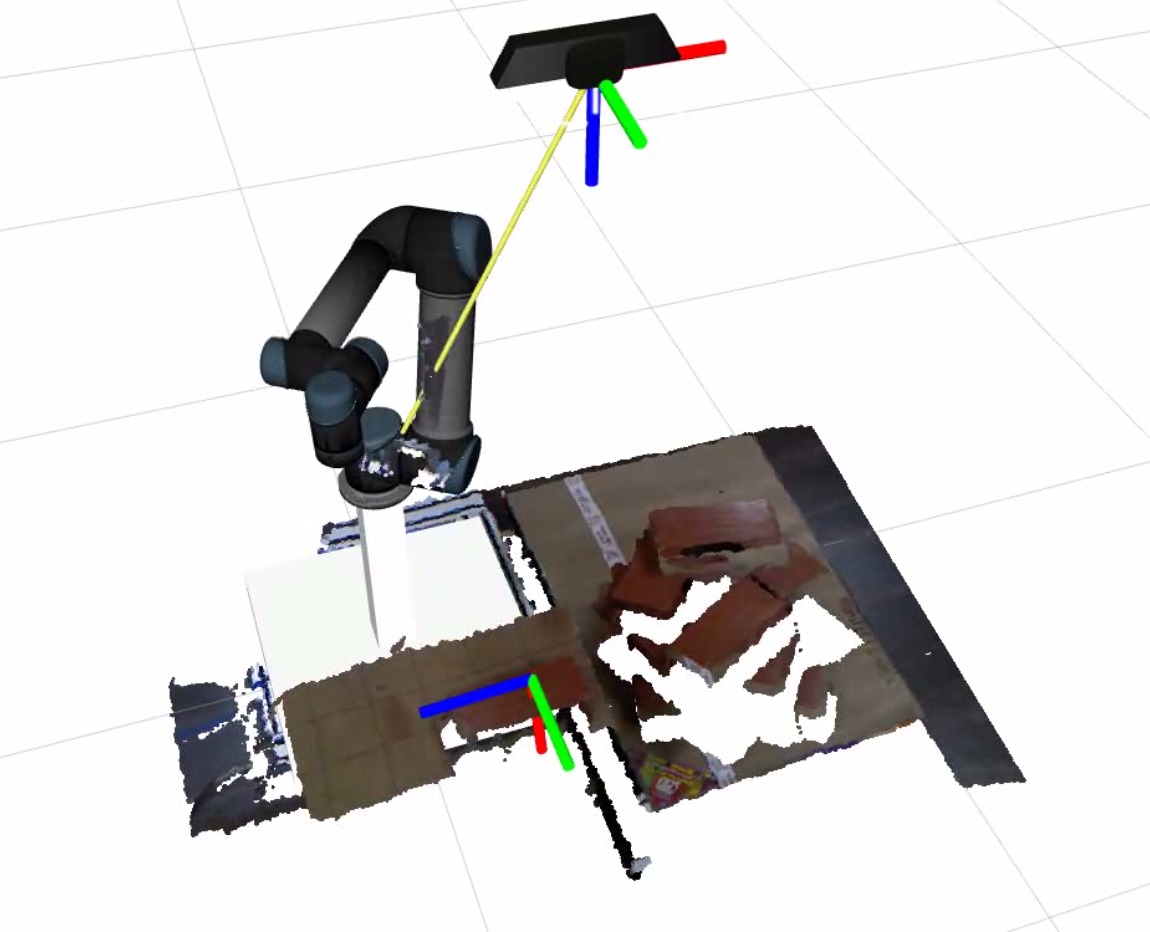}
                \caption{Setup}
                \label{fig:gull}
        \end{subfigure}%
        \begin{subfigure}[b]{0.2\textwidth}
                \centering
                \includegraphics[scale=0.095]{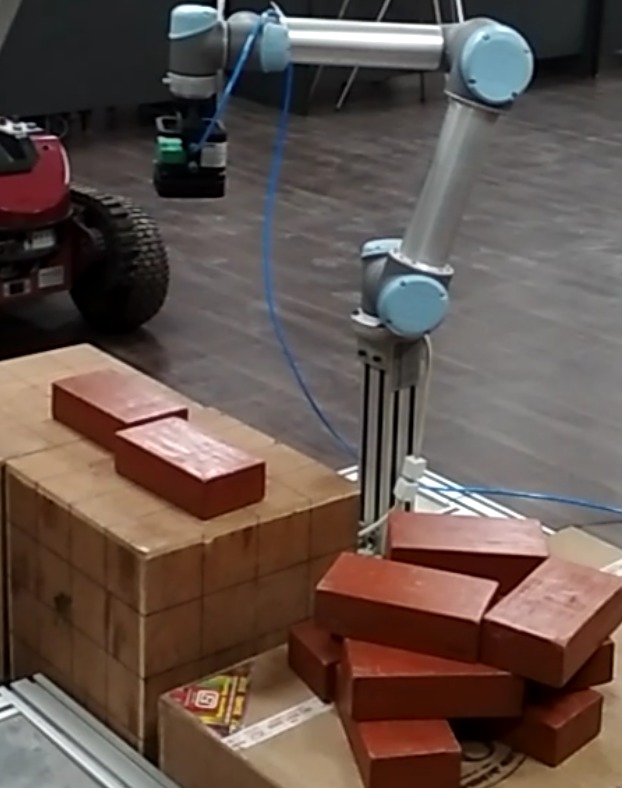}
                \caption{Visual Processing}
                \label{fig:gull}
        \end{subfigure}%
        \begin{subfigure}[b]{0.2\textwidth}
                \centering
                \includegraphics[scale=0.1]{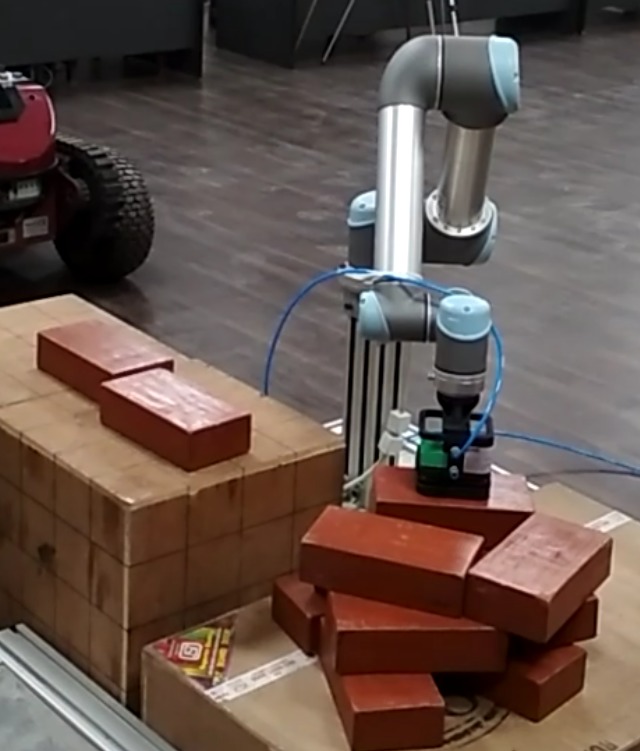}
                \caption{Grasp}
                \label{fig:gull2}
        \end{subfigure}%
        \begin{subfigure}[b]{0.2\textwidth}
                \centering
                \includegraphics[scale=0.095]{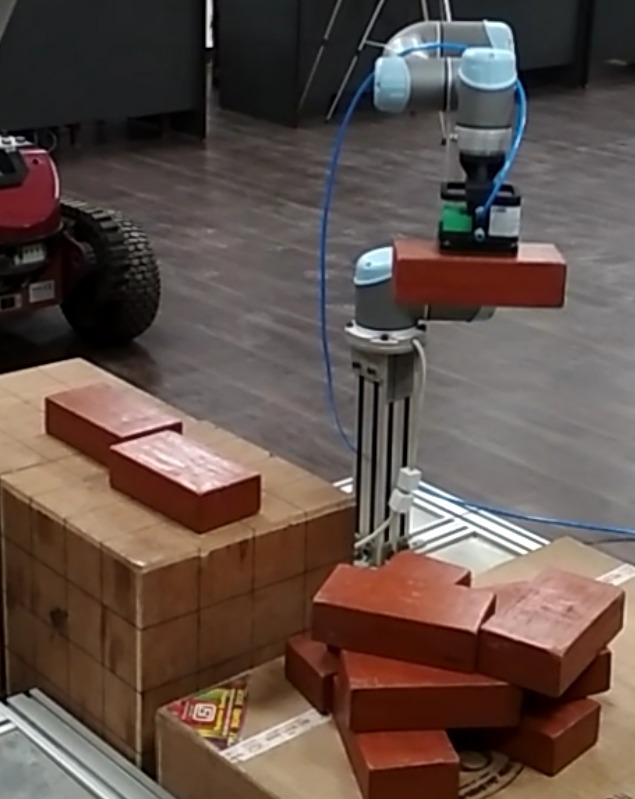}
                \caption{Retrieve}
                \label{fig:tiger}
        \end{subfigure}%
        \begin{subfigure}[b]{0.2\textwidth}
                \centering
                \includegraphics[scale=0.1]{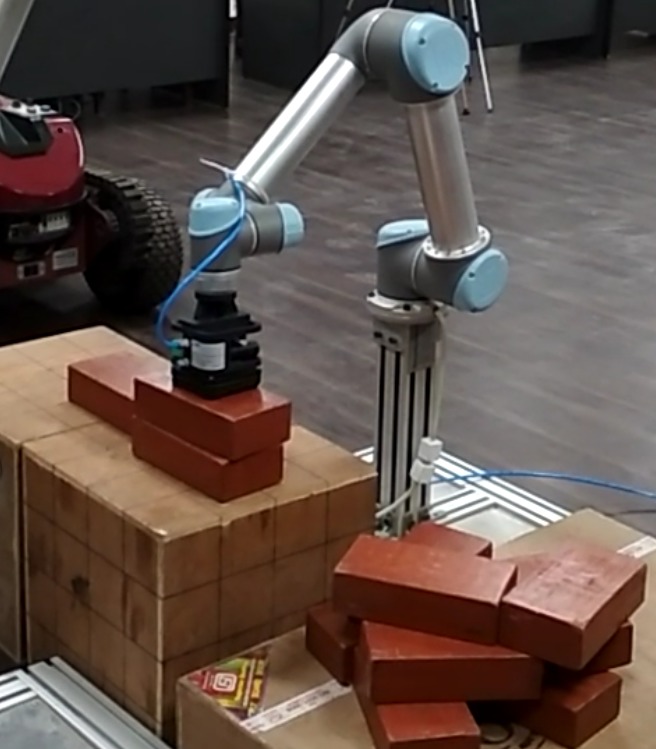}
                \caption{Place}
                \label{fig:mouse}
        \end{subfigure}
        \caption{Experimental setup and execution steps during robotic manipulation}\label{fig:full_system}
        \vspace{-4mm}
\end{figure*}

\subsection{Experimental Setup}
Our robot platform setup is shown in Fig. \ref{fig:full_system}. It consists of a UR5 robot manipulator with its controller box (internal computer) and a host PC (external computer). The UR5 robot manipulator is a 6-DoF robot arm designed to work safely alongside humans. The low-level robot controller is a program running on UR5’s internal computer, broadcasting robot arm data, receiving and interpreting the commands, and controlling the arm accordingly. There are several options for communicating with the robot low-level controller, for example, teach pendant or opening a TCP socket (C++/Python) on a host computer. Our vision hardware consists of an RGB-D Microsoft Kinect sensor mounted on top of the workspace. Point cloud processing is done in the PCL library, and ROS drivers are used for communication between the sensor and the manipulator. A suction gripper is mounted at the end effector of the manipulator.

\subsection{Motion Control Module}
The perception module passes the 6D pose of the object to the main motion control node running on the host PC. The main controller then interacts with low level Robot Controller to execute the  motion. A pick and place operation for an object consists of four actions: \textit{i}) approach the object, \textit{ii}) grasp, \textit{iii}) retrieve  and \textit{iv}) place the object. 
The overall trajectory has five way-points: an initial-point ($I$), a mid-point ($M$), a goal-point ($G$), a retrieval point ($R$), and a final destination point ($F$). A final path is generated which passes through all of the above points.
\par
\textbf{\textit{i) Approach:}}
This part of the trajectory connects $I$ and $M$. The point $I$ is obtained by forward kinematics on current joint state of the robot and $M$ is an intermediate point defined by $M= G + d~\hat{z}$, where $G$ is the goal-point provided by the pose-estimation process and $\hat{z}$ is the unit vector in vertical up direction at $G$. The value of $d$ is selected subjectively after several pick-place trial runs. 
\par
\textbf{\textit{ii) Grasp:}}
As the robot's end effector reaches $M$, the end-effector align according to the brick pose. This section of the trajectory is executed between $M$ and $G$ with directed straight line trajectory of the end effector.
\par
\textbf{\textit{iii) Retrieve:}}
Retrieval is the process of bringing the grasped object out of the clutter. This is executed from $G$ to $R$. The retrieval path is traversed vertically up straight line trajectory. $R$ is given by $R= G + h~\hat{z}$, where $G$ is the goal-point and $\hat{z}$ is the unit vector in vertical up direction at $G$. The value of $h$ is selected to ensure the clearance of the lifted object from the clutter.
\par
\textbf{\textit{iii) Place:}}
Placing is the process of dropping the grasped object at the destination place. This is executed from $R$ to $F$. Fig. \ref{fig:full_system} shows the execution steps during robotic manipulation.

\subsection{Experimental Validation}
\begin{figure*}[t!]
  \centering
      \begin{subfigure}[b]{0.25\linewidth}
    \includegraphics[width=0.9\linewidth]{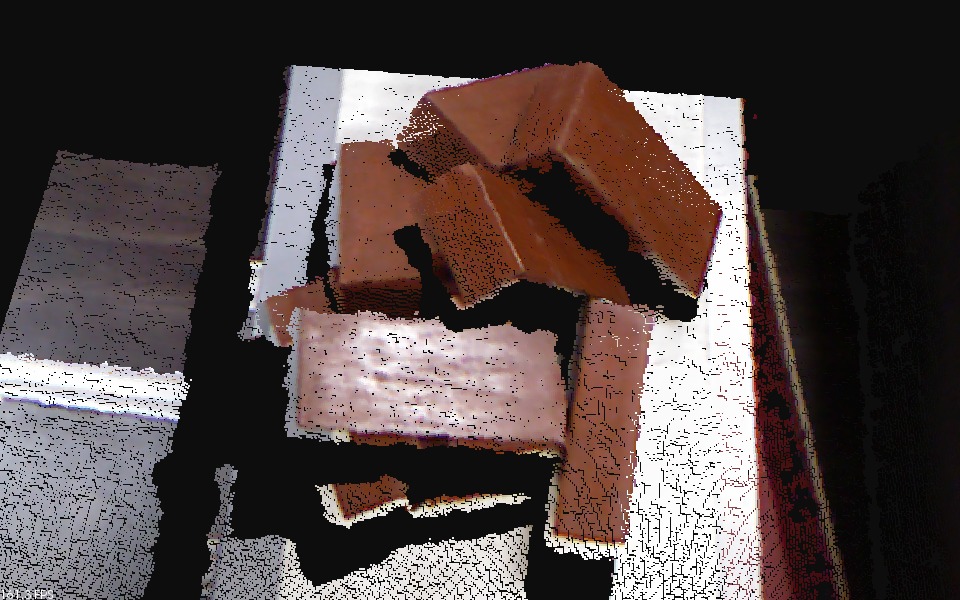}
    \caption{Input}
  \end{subfigure}%
  \begin{subfigure}[b]{0.25\linewidth}
    \includegraphics[width=0.9\linewidth]{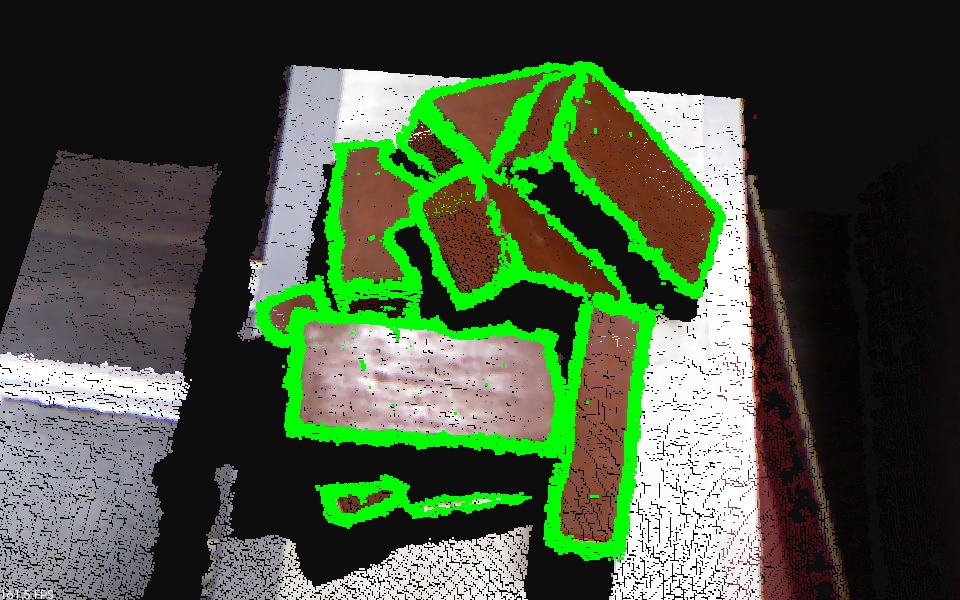}
    \caption{Edges points}
  \end{subfigure}%
  \begin{subfigure}[b]{0.25\linewidth}
    \includegraphics[width=0.9\linewidth]{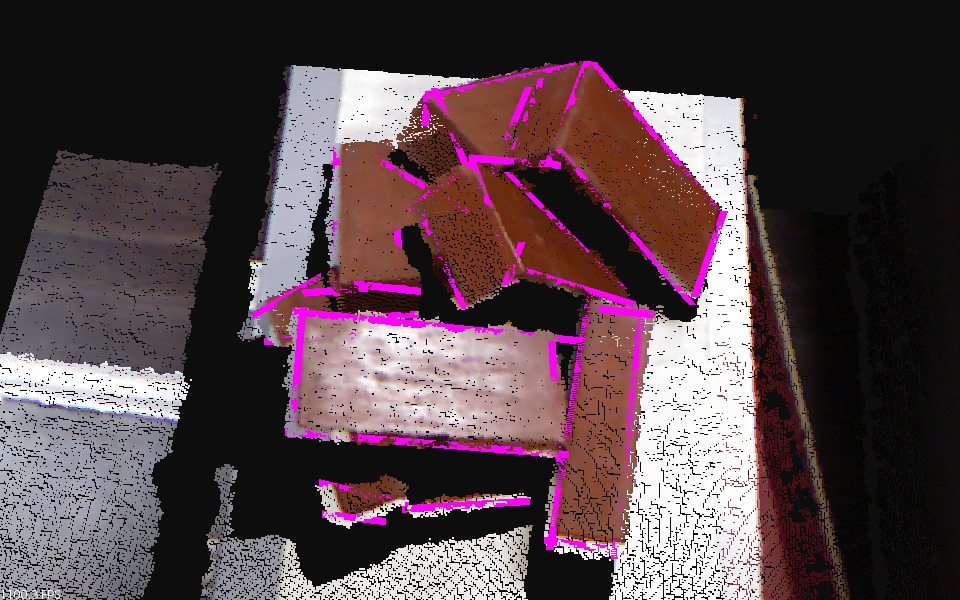}
     \caption{All Edges}
  \end{subfigure}%
    \begin{subfigure}[b]{0.25\linewidth}
    \includegraphics[width=0.9\linewidth]{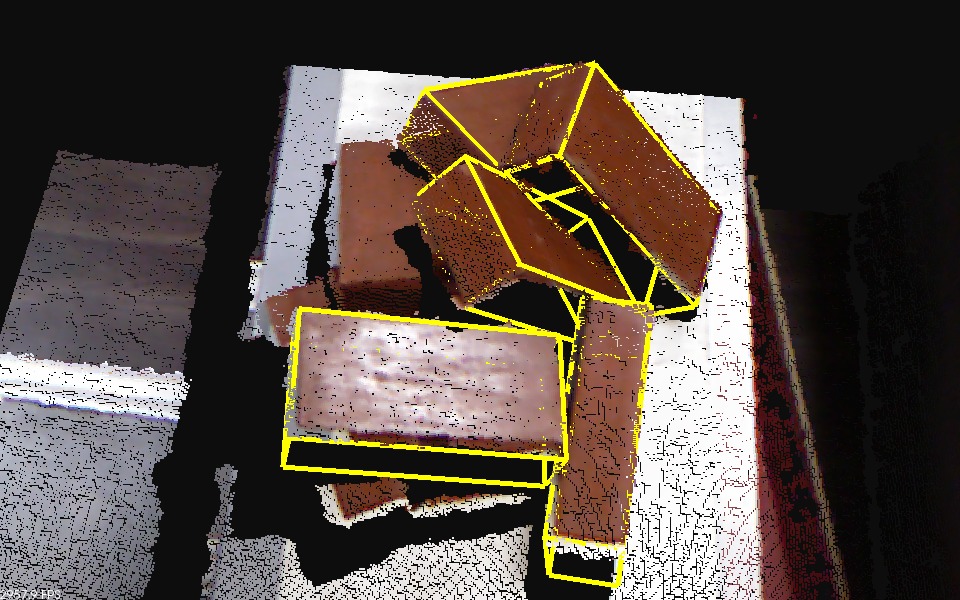}
    \caption{Model Fitting}
  \end{subfigure}
  \caption{Clutter bricks experiment}
  \label{fig:clutter_exp}
\end{figure*}

 We explore two experimental settings: when bricks are presented to the robot in isolation and when bricks are presented in a dense cluttered scenario as shown in Fig. \ref{fig:no_clutter_exp} and Fig. \ref{fig:clutter_exp} respectively.

\subsubsection{Objects present in the isolation}
In the first experiment we measure the dimension of bricks and place the bricks separately in front of the 3D sensor.

\begin{figure*}[t!]
  \centering
      \begin{subfigure}[b]{0.25\linewidth}
    \includegraphics[width=0.9\linewidth]{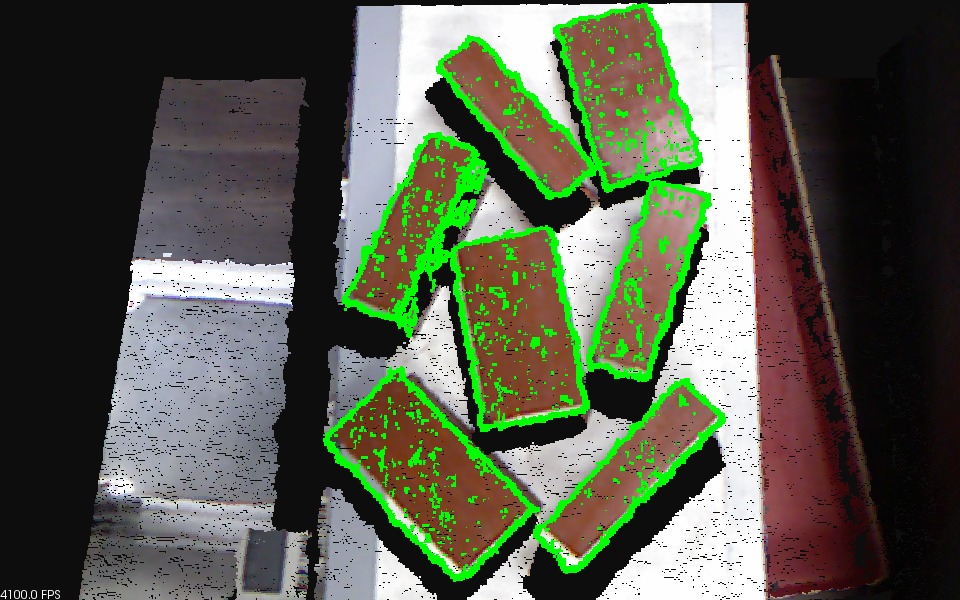}
    \caption{$r_s = 0.01$}
  \end{subfigure}%
  \begin{subfigure}[b]{0.25\linewidth}
    \includegraphics[width=0.9\linewidth]{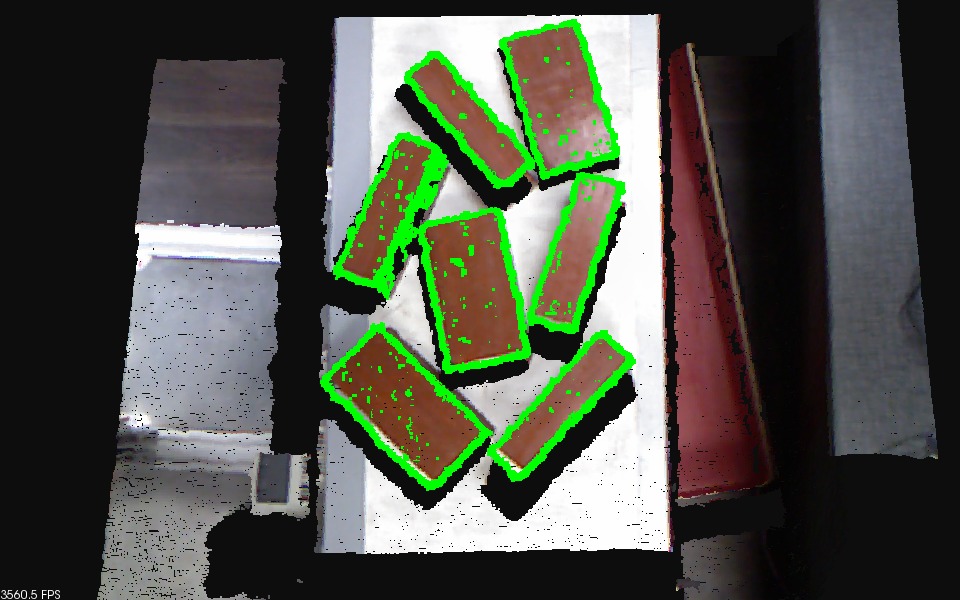}
    \caption{$r_s = 0.015$}
  \end{subfigure}%
  \begin{subfigure}[b]{0.25\linewidth}
    \includegraphics[width=0.9\linewidth]{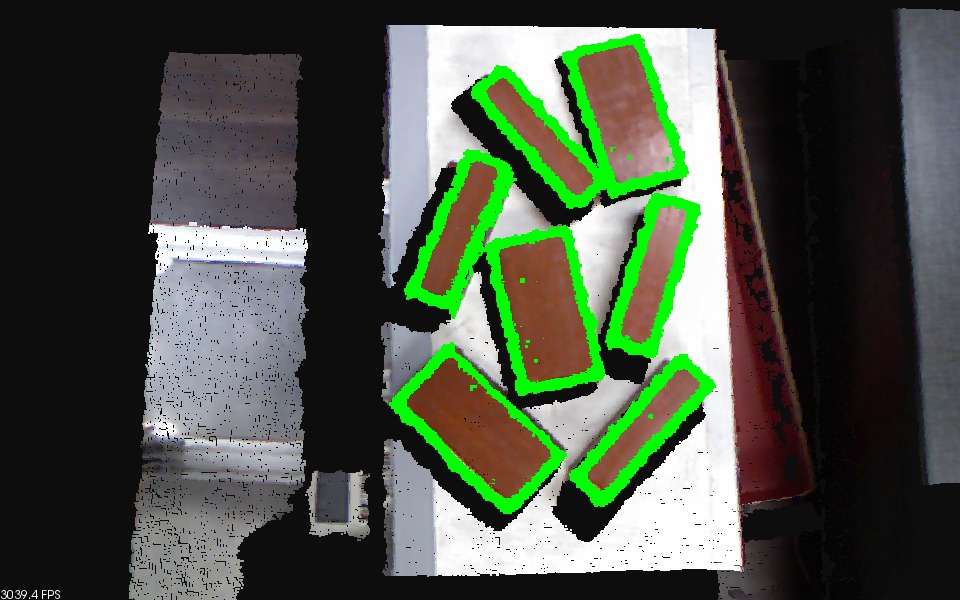}
    \caption{$r_s = 0.025$}
  \end{subfigure}%
    \begin{subfigure}[b]{0.25\linewidth}
    \includegraphics[width=0.9\linewidth]{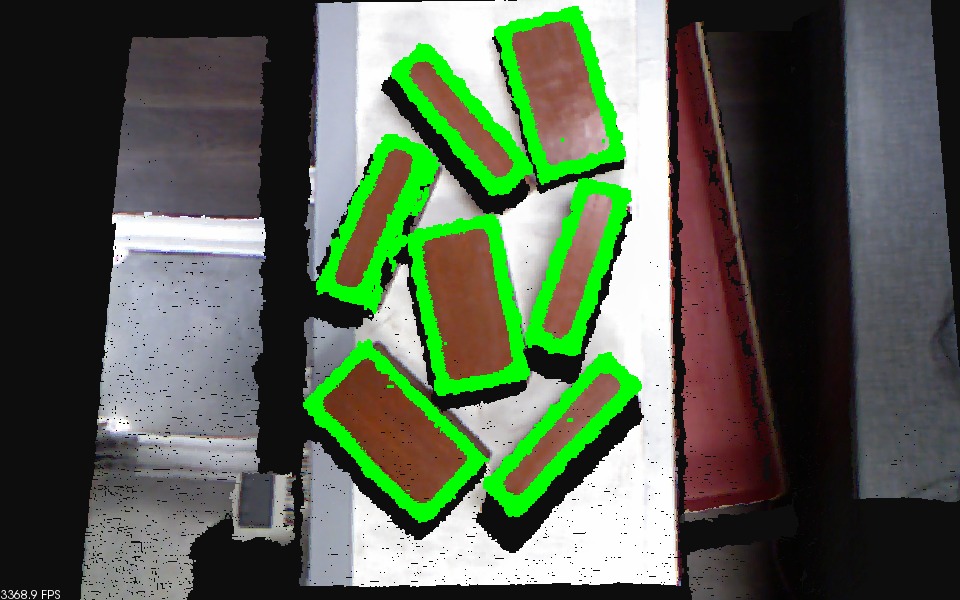}
    \caption{$r_s = 0.030$}
  \end{subfigure}
  \caption{Effect of $r_s$ at $t_h=0.35$}
  \label{fig:r_value_exp}
\end{figure*}

\textbf{Edge points:} For extracting edges, we perform several experiments for selecting the optimum value of $r_s$ and $t_h$. We initiate with $r_s = 0.005m$ and increment it with a step size of $0.001m$. We record the computation time at each $r_s$ (Table \ref{r_table}). Based on quality of edges (Fig. \ref{fig:r_value_exp}) and computation time we select the $r_s$. Based on several trials we found that for extracting edge points, $r_s= 0.02m$ and $t_h = 0.35$ are the optimal values.

\begin{table}[!h]
    \fontsize{7pt}{11pt}\selectfont
	\centering
	\caption{Computation time (sec) at various $r_s$ }
	\label{r_table}
	\begin{tabular}{|c|c|c|c|c|}
	    \hline
		$r_s = $ 0.010 & $r_s = $ 0.015 & $r_s = $ 0.020 & $r_s = $ 0.025 & $r_s =$ 0.030 \\
	    \hline
		0.66538 & 1.29415 & 2.13939 & 3.16722 & 4.40235  \\
		\hline
	\end{tabular}
\end{table}

\begin{figure*}[h!]
  \centering
      \begin{subfigure}[b]{0.25\linewidth}
    \includegraphics[width=0.9\linewidth]{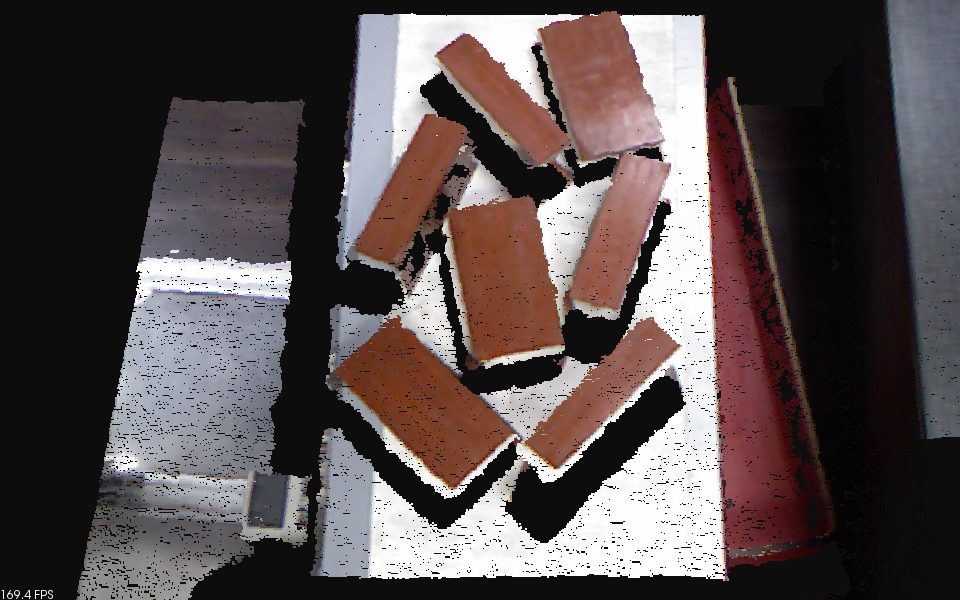}
    \caption{Raw point cloud}
  \end{subfigure}%
  \begin{subfigure}[b]{0.25\linewidth}
    \includegraphics[width=0.9\linewidth]{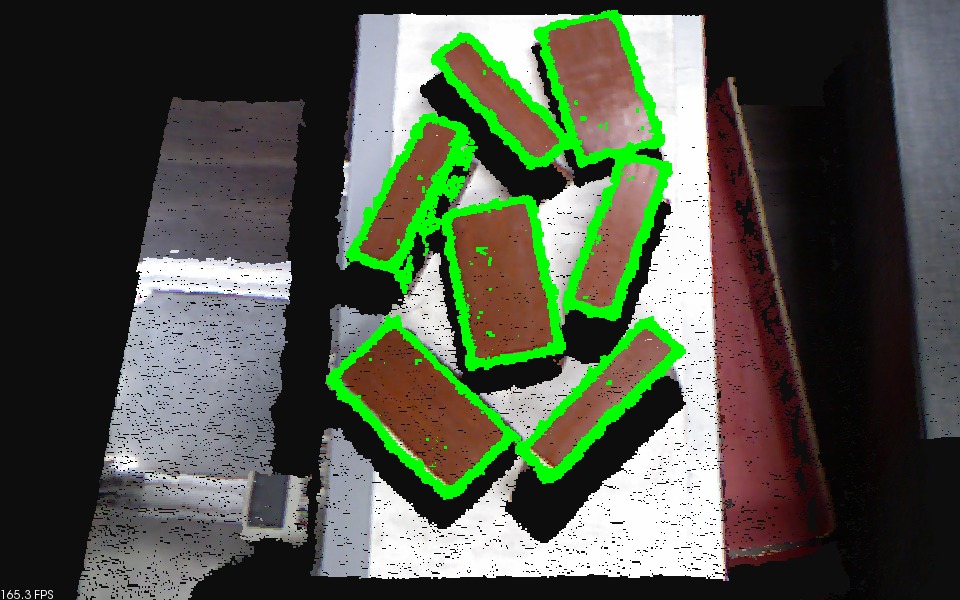}
    \caption{Edges Point}
  \end{subfigure}%
  \begin{subfigure}[b]{0.25\linewidth}
    \includegraphics[width=0.9\linewidth]{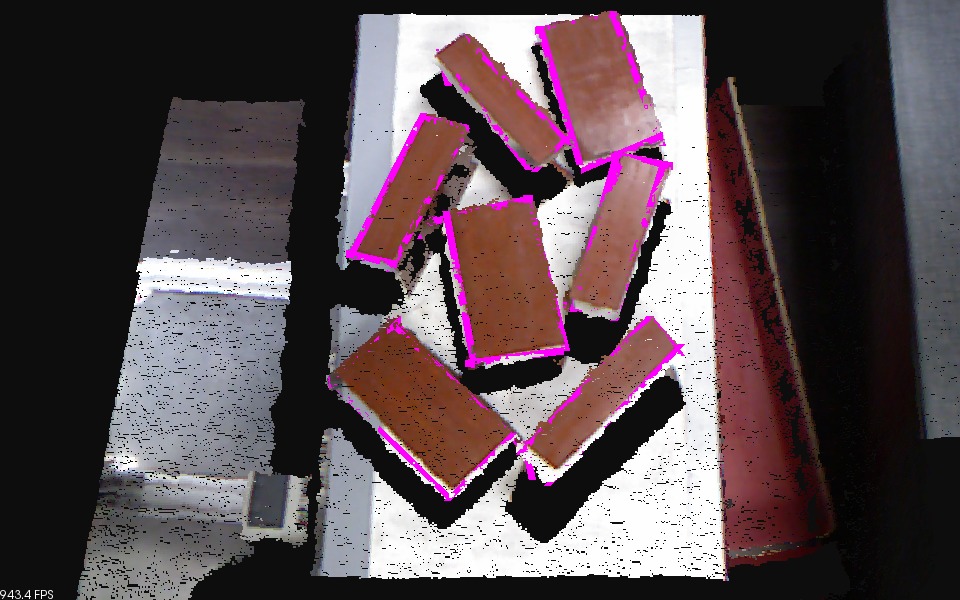}
    \caption{All Edges}
  \end{subfigure}%
    \begin{subfigure}[b]{0.25\linewidth}
    \includegraphics[width=0.9\linewidth]{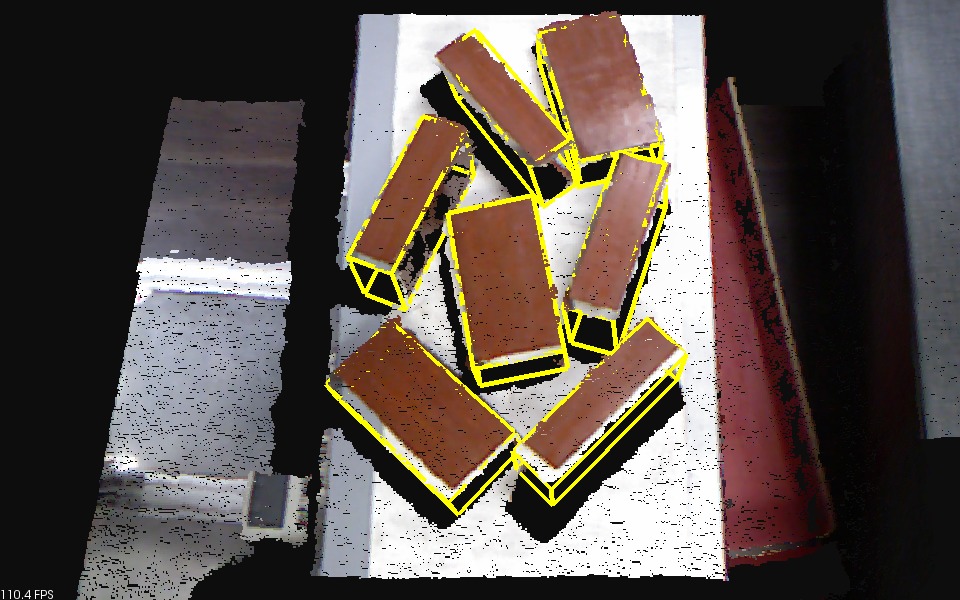}
    \caption{Model Fitting}
  \end{subfigure}
  \caption{Isolated bricks experiment}
  \label{fig:no_clutter_exp}
\end{figure*}

\textbf{Lines and corner points:} For finding the equation of lines we apply \textit{RANSAC} method with threshold value of $0.01m$ and for finding the corner point, we compute the distance between two extremities of edges and if distance is $ < 0.01m$, then corner point is the average of two extreme points. With these parameters, we find the edges and corner points which is shown in Fig. \ref{fig:no_clutter_exp} and computation time for each step is shown in Table \ref{Computation_time_table}.

\begin{table}[ht]
    \fontsize{7pt}{11pt}\selectfont
	\centering
	\caption{Computation time (sec) at each step}
	\label{Computation_time_table}
		\begin{tabular}{|c|c|c|c|c|}
		\hline 
	    \textbf{} & Edges Point  & All edges &  Model Fitting  & Total Time \\
	    \hline 
	    \textbf{Exp }$\mathbf{1}$ & 2.13939  & 2.0854  &  0.07573  & 4.36192 \\
	    \hline 
	    \textbf{Exp }$\mathbf{2}$ & 2.13056  & 5.26684 &  0.21726  & 7.61466 \\
		\hline
	\end{tabular}
\end{table}

\subsubsection{Clutter of bricks}
In the second experiment we place the brick in clutter as shown in Fig. \ref{fig:clutter_exp}. For edge points and lines, we use the same parameter values which were used in experiment 1. Result of each step is shown in Fig. \ref{fig:clutter_exp} and computation time for each step is shown in Table \ref{Computation_time_table}. With each successful grasp of the brick, computation time will decreases, because the number of points which needs to be processed will decrease. Above experiment is performed on a system with an i7 processor having a clock speed of 3.5GHz and 8GB RAM.

\subsection{Performance on different objects}

\begin{figure}[ht]
  \centering
      \begin{subfigure}[b]{0.4\linewidth}
    \includegraphics[width=\linewidth]{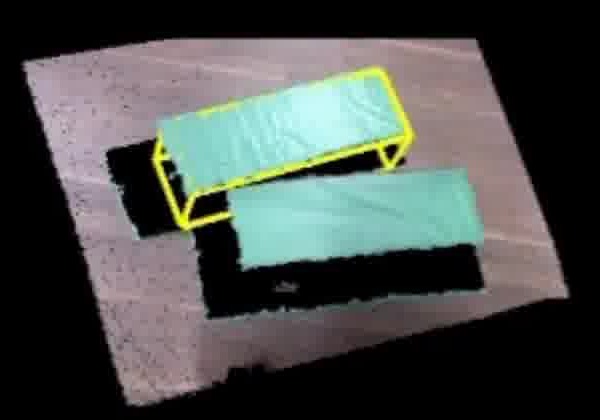}
  \end{subfigure}
  \begin{subfigure}[b]{0.4\linewidth}
    \includegraphics[width=\linewidth]{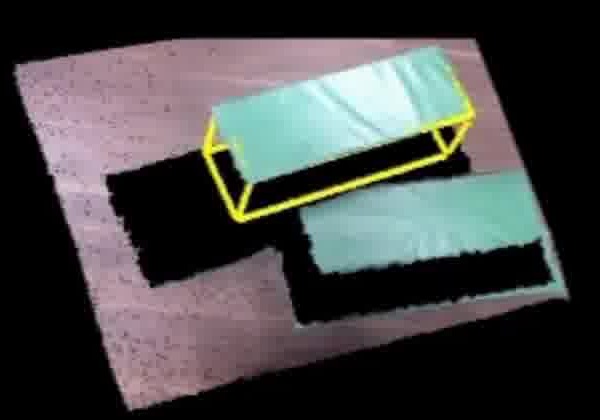}
  \end{subfigure}
        \begin{subfigure}[b]{0.4\linewidth}
    \includegraphics[width=\linewidth]{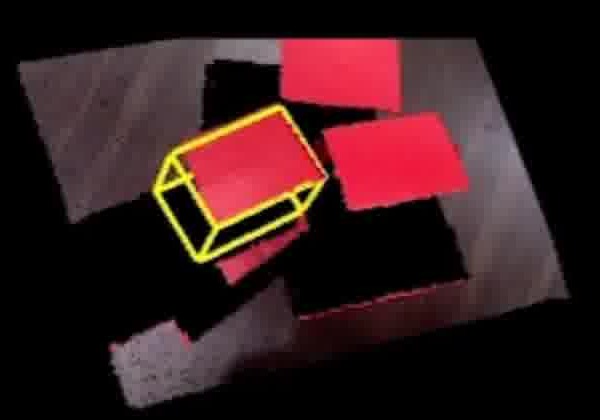}
  \end{subfigure}
  \begin{subfigure}[b]{0.4\linewidth}
    \includegraphics[width=\linewidth]{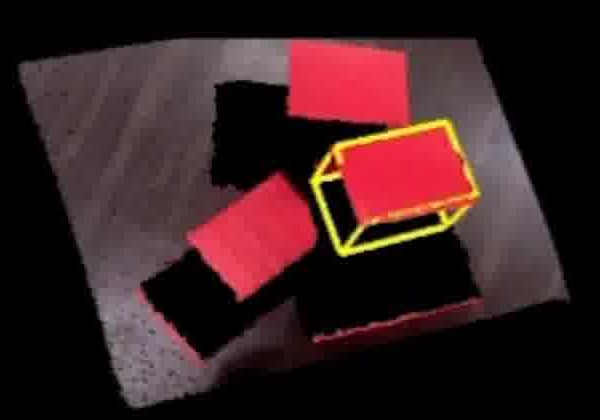}
  \end{subfigure}
   \caption{Pose predictions for different objects}
  \label{fig:diff_obj}
\end{figure}

As mentioned earlier, our method can be easily applied for the other objects with known dimensions. Hence we tested our algorithm on two different objects with different dimensions. Fig. \ref{fig:diff_obj} shows the output of our method on two different objects (only dimensions are known in advance). A single pose has been visualized in Fig. \ref{fig:diff_obj} to avoid a mess in the images. From Fig. \ref{fig:diff_obj}, we can claim that our method can easily be deployed for other objects.

\subsection{Performance Analysis}
\begin{figure}[ht]
  \centering
      \begin{subfigure}[b]{0.4\linewidth}
    \includegraphics[width=\linewidth]{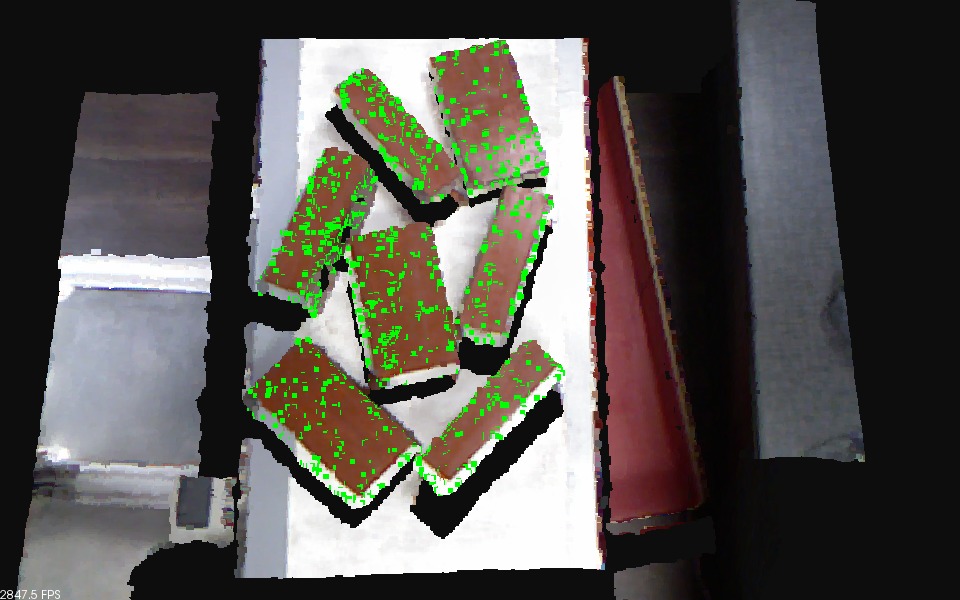}
    \caption{$k=4$}
  \end{subfigure}
  \begin{subfigure}[b]{0.4\linewidth}
    \includegraphics[width=\linewidth]{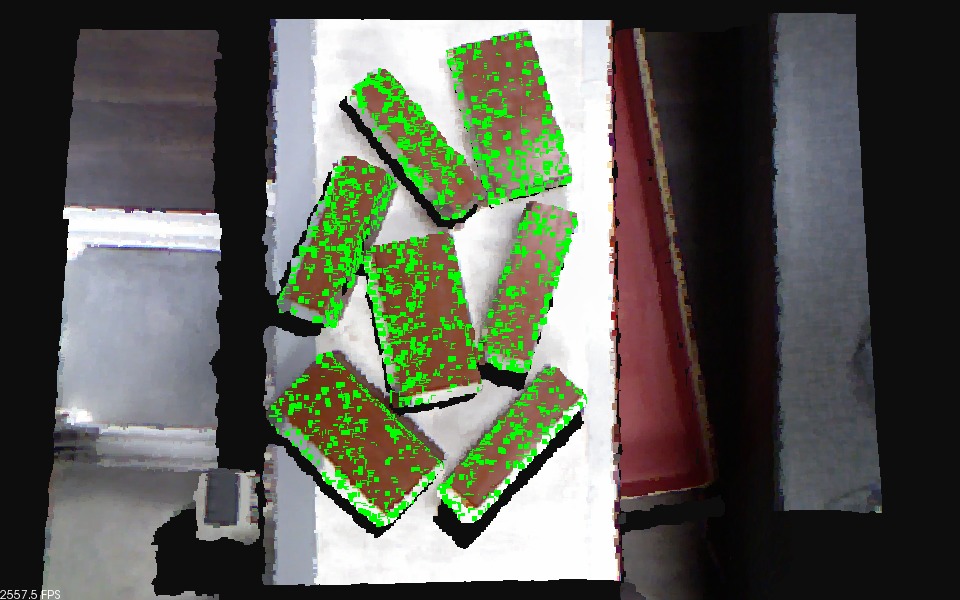}
    \caption{$k=5$}
  \end{subfigure}
        \begin{subfigure}[b]{0.4\linewidth}
    \includegraphics[width=\linewidth]{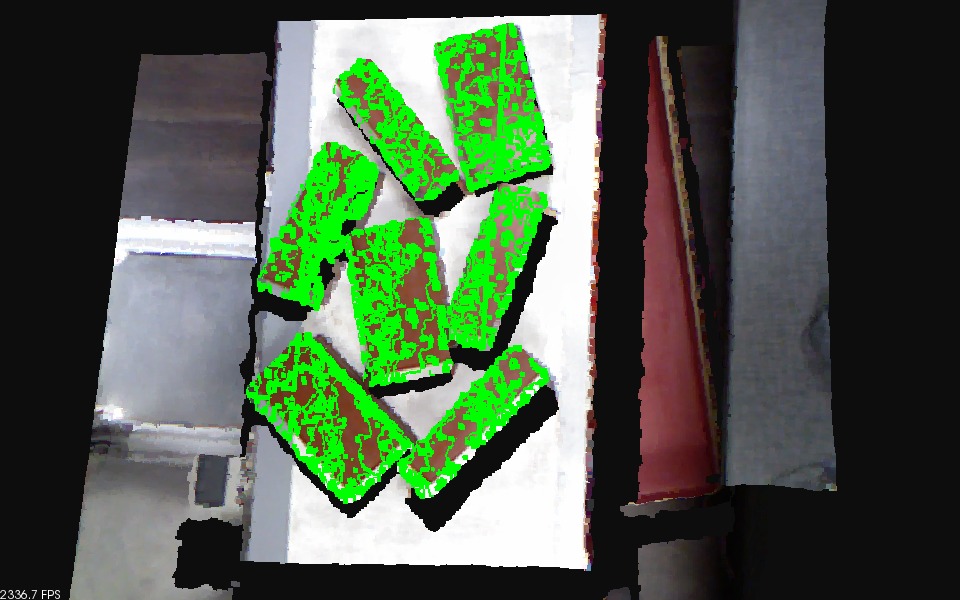}
    \caption{$k=10$}
  \end{subfigure}
  \begin{subfigure}[b]{0.4\linewidth}
    \includegraphics[width=\linewidth]{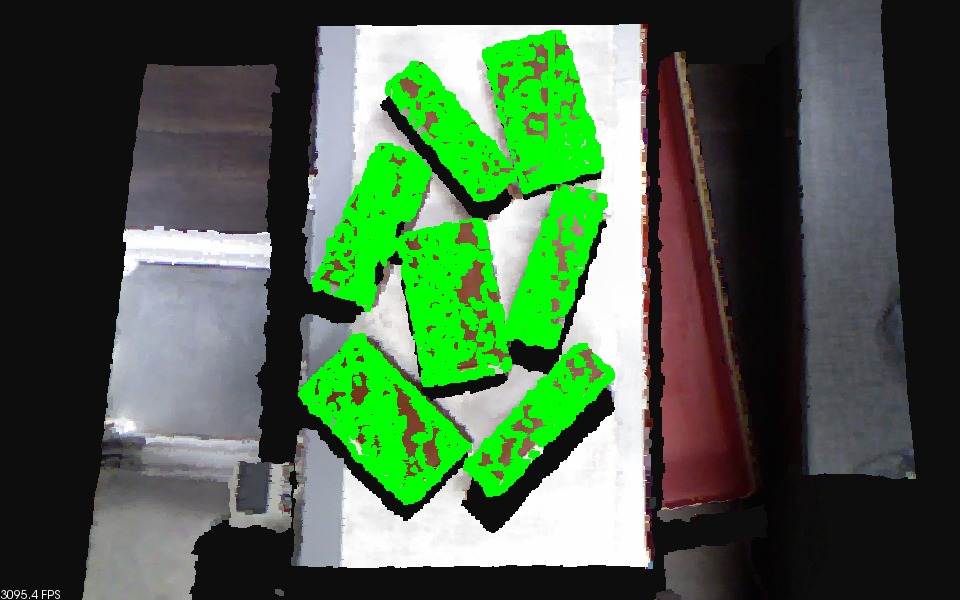}
    \caption{$k=30$}
  \end{subfigure}
   \caption{Edges from covariance matrix based method}
  \label{fig:analysis}
\end{figure}

To demonstrate the efficacy of our proposed edge extraction method from unorganized point cloud, we compare the results from the method \cite{bazazian2015fast} for edge extraction. Their method estimates the sharp features by analysing the eigenvalues of the covariance matrix which is defined by each point\textquotesingle s $k$-nearest neighbors. We apply their method\footnote{https://github.com/denabazazian/} on raw point cloud data with $k$ vary from $4$ to $30$ as shown in Fig. \ref{fig:analysis}. The results of the method \cite{bazazian2015fast} is unsatisfactory on the same data set where our method perform very well. It is because of the inherent noise in the sensor, points on a flat surface has large variations. Thus all eigenvalues of the covariance matrix will be large, hence predicts the sharp edges even at the flat surface.

\section{Conclusion}
Novel edge and corner detection algorithm for unorganized point clouds was proposed and tested on generic objects like a coffee mug, dragon, bunny, and clutter of random objects. The algorithm is used for 6D pose estimation of known objects in clutter for robotic pick and place applications. The proposed technique is tested on two warehouse scenarios, when objects are placed distinctly and when objects are placed in a dense clutter. Results of each scenario is reported in the paper along with the computation time at each step. To demonstrate the efficacy of the edge extraction technique, we compared it with the covariance matrix based solution for 3D edge extractions from unorganized point cloud in a real scenario and report better performance. The overall approach is tested in a warehouse application where a real UR5 robot manipulator is used for robotic pick and place operations.

\bibliographystyle{apalike}
{\small
\bibliography{citations.bib}}

\end{document}